\documentclass[pdflatex,sn-mathphys-ay]{sn-jnl}

\usepackage{graphicx}%
\usepackage{multirow}%
\usepackage{amsmath,amssymb,amsfonts,bm}%
\usepackage{amsthm}%
\usepackage{mathrsfs}%
\usepackage[title]{appendix}%
\usepackage{xcolor}%
\usepackage{textcomp}%
\usepackage{manyfoot}%
\usepackage{booktabs}%
\usepackage{algorithm}%
\usepackage{algorithmicx}%
\usepackage{algpseudocode}%
\usepackage{listings}%
\usepackage{placeins}
\usepackage{enumitem}
\usepackage{url}
\usepackage{fnpct}

\newcommand{\myvect}[1]{\bm{#1}}
\newcommand{\mymat}[1]{\bm{#1}}

\newcommand{\successtag}{\ensuremath{\text{succ}^{\ast}}}

\theoremstyle{thmstyleone}%
\theoremstyle{thmstyletwo}%

\theoremstyle{thmstylethree}%

\newcommand{\emo}{{emotion-like control}}

\newcommand{\EMO}{{Emotion-like Control}}

\begin{document}

\title[Synthetic Emotions \& Consciousness]{Synthetic Emotions and Consciousness: Exploring Architectural Boundaries}

\author{Hermann Borotschnig}







\abstract{
As artificial agents display increasingly sophisticated emotion-like behaviors, frameworks for assessing whether such systems risk instantiating consciousness remain limited.
This contribution asks whether synthetic \emo{} can be implemented while deliberately \emph{excluding} architectural features that major theories associate with \emph{access-like} consciousness.
We propose architectural principles (\textbf{A1}--\textbf{A8}) for a hierarchical, dual-source implementation in which (i) immediate needs generate motivational signals and (ii) episodic memory provides affective guidance from similar past situations; the two sources converge to modulate action selection.
To operationalize consciousness-related risk, we distill predictions from major theories into four engineering risk-reduction constraints: (\textbf{R1}) no content-general, workspace-like global broadcast, (\textbf{R2}) no metarepresentation, (\textbf{R3}) no autobiographical consolidation, and (\textbf{R4}) bounded learning.
We address three questions:
(\textbf{Q1}) Can \emo{} satisfy R1--R4? We present a concrete architecture as an existence proof.
(\textbf{Q2}) Can the architecture be extended without introducing access-enabling features? We identify stable modifications that preserve compliance.
(\textbf{Q3}) Can we trace graded paths that plausibly increase access risk? We map gradual transitions that progressively violate the constraints.
While we cannot resolve questions about consciousness, our contribution operates at three levels: on the engineering side, we present a modular, biologically motivated control architecture; on the theoretical side, we propose a control model of emotions and a methodological template for converting consciousness-related questions into auditable architectural tests; on the safety side, we sketch preliminary audit indicators that may inform future governance frameworks.
The architecture functions independently as an emotion-like controller, while the risk-reduction criteria may extend to other AI systems.
We offer no definitive answers about phenomenology, but we provide starting points for navigating uncertainty---tools intended to evolve as understanding advances.
}


\maketitle

\section{Introduction}\label{sec:intro}

Artificial systems now exhibit flexible decision-making, memory, and context-sensitive behavior at a scale that brings long-standing questions about machine consciousness, emotion, and moral status into practical focus~\citep{seth2025conscious,butlin2025identifying}.
This contribution asks a focused question of public and policy relevance---and of philosophical interest: \emph{Can artificial systems implement \emo{} as a practical control layer while deliberately avoiding architectural features that major theories associate with access-like consciousness?}

Artificial architectures offer a tractable domain for probing such questions.
We argue that an agent need not self-ascribe feelings in order to operate via affect-like mechanisms.
In our operational sense, \emo{} constitutes a specific family of heuristics for resource-bounded action selection.
More concretely, it is a biologically motivated control architecture (A1--A8) in which need-based appraisal and episodic affective memory jointly generate affective states that bias action selection.
Affect is not peripheral in this design: affective variables pervade need appraisal, memory encoding, retrieval, fusion, and decision-making, such that the system can neither be understood nor coherently specified without them.

\paragraph{The double risk.}
The relationship between such systems and access-like consciousness carries practical urgency because the risks are asymmetric.
On the one hand, emotionally expressive but non-conscious systems may deceive users into forming pseudo-relationships with simulacra, enabling manipulation and psychological harm without any moral patient being present.
On the other hand, systems may gain functional sophistication that raises consciousness concerns---particularly when emotion-like mechanisms guide decision-making via standard machine-learning (ML) practices such as value tagging, episodic control~\citep{pritzel2017}, and deep reinforcement learning pipelines~\citep{Mnih2015NatureDQN,gershman2017reinforcement}.
The opacity of consciousness detection compounds both risks: we cannot easily verify whether an expressive system lacks experience (protecting users) or whether a functional system has acquired it (protecting the system).
Understanding couplings between emotion-like control and consciousness-relevant architecture is therefore important for responsible development.

\begin{table*}[h!]
\centering
\scalebox{1.0}{
\begin{tabular}{@{}p{0.28\linewidth}p{0.68\linewidth}@{}}
\toprule
\multicolumn{2}{l}{\textbf{Core Concepts}} \\
\midrule
Emotion-like control & Hierarchical, dual-source architecture combining need appraisals and episodic memory to bias policy selection and action instantiation (see A1--A8 below). Not a claim about phenomenology. \\
Separation witness & A concrete implementation of \emo{} A1--A8 that satisfies R1--R4, showing that affective control can be realized without introducing the access-enabling mechanisms targeted by our proxy (not a proof of non-consciousness). \\
Plausibly access-excluding & Systems satisfying constraints R1--R4; the target conservative design region under our multi-family proxy. \\
\midrule
\multicolumn{2}{l}{\textbf{Architectural Components (A1--A8)} for control loop $\mathcal{S}$} \\
\midrule
A1 Hierarchy & Two-level control: behavioral modes or policies are selected by affect, then instantiated to concrete actions. Observed situations are mapped to categorical abstractions. \\
A2 Need appraisal & Transforms homeostatic discrepancies into drives, affect, and policy preferences. \\
A3 Episodic memory & Stores situation-categories with affect tags and policy hints; enables similarity-based retrieval. \\
A4 Integration & Combines need-based and memory-based signals into unified affect and policy preferences. \\
A5 Policy instantiation & Maps selected policy to preferences for concrete situation-dependent actions. \\
A6 Action execution & Performs instantiated action and observes state change from pre-act state to post-act situation. \\
A7 Post-action reappraisal & Observes outcome and generates post-action affect. Evaluates outcome based on need and affect changes. \\
A8 Episode storage & Stores categorical abstraction of pre-act situation, pre-/post-act affect, policy preferences, and post-action success. \\
\midrule
\multicolumn{2}{l}{\textbf{Design Constraints (R1--R4)}} \\
\midrule
R1~No~shared workspace & No content-general workspace/hub/router that makes internal representations flexibly reusable across heterogeneous subsystems (no workspace-like broadcast); allows narrow, typed control signals. \\
R2~No~metarepresentation & No higher-order representations about having internal states. \\
R3~No autobio. consolid. & No construction, persistence, export, or re-ingestion of identity/time-anchored cross-episode summary state (no autobiographical ``glue'' or consolidation). \\
R4~Bounded~learning & Credit assignment stays within module boundaries. \\
\midrule
\multicolumn{2}{l}{\textbf{Research Questions (Q1--Q3)}} \\
\midrule
Q1 Existence & Can \emo{} A1--A8 also satisfy R1--R4? \\
Q2 Stability & Are there design modifications preserving witness status? \\
Q3 Gradualism & Can we trace graded paths plausibly increasing access risk? \\
\midrule
\multicolumn{2}{l}{\textbf{Consciousness Framework}} \\
\midrule
Access-like consciousness & Global availability, higher-order access, integration. \\
Phenomenal consciousness & Felt, valenced experience (normative anchor, not analyzed). \\
Multi-family proxy & GWT/GNW, HOT-like theories used to motivate R1--R4 constraints (see~\S\,\ref{sec:GWTHOT}). \\
\bottomrule
\end{tabular}
}
\caption{Terminology at a Glance}
\label{tab:Terminology}
\end{table*}

\paragraph{General Assumptions.}
We proceed under the following assumptions, fixing scope and interpretation (see~Table~\ref{tab:Terminology} for terminology):

\begin{itemize}[leftmargin=1.25em,itemsep=0.25em]

\item \textbf{Functionalism.} We adopt a functionalist stance: roles and causal organization of processing, rather than substrate, ground our analysis. Our claims concern architecture and information flow rather than implementation specifics.

\item \textbf{Control role of emotion.} We treat emotions as control processes biasing retrieval, evaluation, and action selection under uncertainty and resource limits~\citep{simon1967motivational}. This is compatible with appraisal traditions that link core affect to behavior~\citep{scherer2001appraisal,frijda1986emotions,Russell2003CoreAffect}, with affective-neuroscience views emphasizing subcortical motivational circuitry~\citep{panksepp1998affective,panksepp2012archaeology,LeDoux2003CMN}, and with computational approaches treating value representations as goals for instrumental action~\citep{rolls2014emotionbook,rolls2014emotionprecis}. We explore whether emotions can function as simple heuristic shortcuts rather than requiring complex integrative machinery.

\item \textbf{Architecture over taxonomies.} We remain agnostic about full emotion taxonomies.\footnote{Rather than adjudicating between basic-emotion~\citep{ekman1992argument}, dimensional~\citep{Russell2003CoreAffect}, appraisal~\citep{scherer2001appraisal}, or constructionist~\citep{barrett2017construction} frameworks, we focus on emotion's control function: how affective states bias action selection under uncertainty.}
For concreteness, we use a minimal descriptive set (valence, arousal, drive) as control variables within an architectural scaffold, consistent with core-affect and appraisal accounts~\citep{Russell2003CoreAffect,scherer2001appraisal}.

\item \textbf{First-order vs.\ higher-order.} We distinguish first-order state tracking (e.g., needs/appraisals) from higher-order metarepresentations (representations \emph{about} those states). This distinction matters for the separation question.
\end{itemize}

\paragraph{Scope and terminology.}
We use \emph{access-like consciousness} operationally, following the standard access/phenomenal distinction~\citep{block1995confusion}: roughly, contents that are globally available for flexible use (workspace-like broadcast), that potentially involve higher-order access, and/or that exhibit strong temporal integration (e.g., ignition, stabilization)~\citep{DehaeneNaccache2001Workspace,Rosenthal2005ConsciousnessMind,lau2011hot}.
By contrast, \emph{phenomenal consciousness} (``what it is like'') may be a \emph{normative anchor} but is not as such our object of positive theorizing here.
We do not attempt to explain qualia or address the hard problem.
Instead, we analyze how concrete design choices (R1--R4) may \emph{raise or lower the risk} that access-enabling features emerge.
Throughout, ``access'' terms are used for operational assessment.
We do not provide complete policy-relevant risk assessments; we provide methodological suggestions that may structure such endeavors, particularly under the assumption that adding access-enabling features weakly increases the probability of phenomenality.

Our stance aims to preserve theoretical neutrality while enabling policy-relevant guidance: we propose to audit access-related \emph{conjuncts} (broadcast$+$ignition; first-order$+$higher-order; information-generation$+$integration) without taking a stand on their metaphysical sufficiency. Our goal is not to decide consciousness in practice, but to understand how design choices may increase or decrease access-consciousness risks. Future ethical recommendations might track \emph{risk} to phenomenal affect, but the present study primarily explores the \emph{possibility} of avoiding access features under constraints.

\paragraph{A reference model for \emo{}.}

We now specify a \emph{hierarchical, dual-source} process as our working notion of ``\emo{}''. The aim is to define emotion-like control architectures that establish an architectural \emph{floor}---a minimal threshold that should nonetheless be rich enough to deserve the label, providing genuine behavioral regulation through need appraisal, episodic retrieval, and affect-modulated selection. To this end, we embed \emo{} into decision-making under uncertainty and resource limits~\citep{simon1967motivational,damasio1999feeling}. The model is biologically and psychologically motivated (see \S\,\ref{sec:biology}) and can be realized in multiple computational styles (see \S\,\ref{sec:A1toA8}). 

By \emph{\emo{}} we thus understand an implementation with the following architectural components and control flow:

\begin{enumerate}[leftmargin=1.5em,itemsep=0.25em,label=\textbf{A\arabic*}]

\item \textbf{Hierarchical organization with categorical abstraction.} The system employs a two-level hierarchy to avoid combinatorial explosion: behavioral policies (such as ``approach'' or ``flee'') are selected based on affective state and episodic hints, then instantiated into actions. Observations are similarly abstracted: infinite sensory variations are translated into situation categories.

\item \textbf{Source 1: Unmet needs and appraisals.} Homeostatic discrepancies generate corresponding (i) drives (graded urgency), (ii) affect (e.g., valence, intensity, arousal), and possibly (iii) initial biases toward specific behavioral policies.

\item \textbf{Source 2: Episodic emotional memory.} Prior episodes (single situations or short sequences) are stored as discrete records of situation-categories with affective tags, providing an alternative source for current emotions and policy hints. Retrieval by situation-category similarity yields (i) contextual affect and (ii) policy suggestions, but not episode content: the controller receives an aggregated ``hunch''---affect and behavioral bias---without access to its source (no ``recollections'' of past episodes\footnote{Our ``episodic memory'' differs from Tulving's~\citep{Tulving1985MemoryConsciousness}, which involves autonoetic consciousness, see~\S\,\ref{sec:GWTHOT}.  Our system is closer to ``episodic-like'' memory~\citep{Clayton1998EpisodicLike} but more implicit still: the controller receives only aggregated affect and policy biases, no details allowing deliberation.}).

\item \textbf{Affective and policy integration.} The two sources converge: need-based and memory-based affects combine into a unified affective state that also influences how policy preferences from both sources are integrated into a set of current policy preferences. This dual integration arbitrates between different sources (current needs and past episodes) and among competing drives.

\item \textbf{Policy instantiation as actions.} The selected policy is mapped to concrete actions through situation-dependent action proposals. These action proposals---whether evolutionarily hardwired or developmentally learned---translate abstract behavioral policies into situation-appropriate actions (e.g., ``flee'' may map to ``move away from threat source'').

\item \textbf{Action execution and outcome observation.} The instantiated action is performed, and the resulting state change is observed.

\item \textbf{Post-action reappraisal.} Success is evaluated based on outcome measures (e.g., changes in need satisfaction), generating post-action affect that reflects the result.

\item \textbf{Episode storage.} The completed episode data is written to memory, enabling future retrieval in similar situations. Episode data includes at least the situation category (primary retrieval key), associated affect values (to contribute to present-moment affect upon retrieval, but potentially also used as secondary keys), indicators of the policies considered/executed, and success.
\end{enumerate}

Within the proposed architectural constraints we remain taxonomy-agnostic with regard to synthetic \emph{affective states and tags}. For explanatory purposes, our examples use a minimal descriptive set in which each \emph{valence} signal is paired with an emotional \emph{intensity}, alongside \emph{drive} (goal-directed urgency from need discrepancy) and \emph{arousal} (activation/vigilance that shifts decisiveness and exploration rate)~\citep{Russell2003CoreAffect}.

When we speak of \emo{}, we mean the above architectural, control-level construct: a hierarchical system where appraisal/drive signals modulate policy selection rather than directly controlling actions, plus episodic write-back for learning. This is \emph{not} a claim about phenomenology and \emph{not} a self-model commitment. Some theorists reserve ``emotion'' for phenomena essentially tied to a self or to experience~\citep{Goldie2002Emotions, Helm2009EvaluativeFeelings}; our target here is narrower and we retain any familiar labels solely as handles. 

Correspondingly, we use ``emotion'' informally for \emo{} (both biological and artificial or synthetic). Our definition should exclude mere mimicry (e.g., a variable arbitrarily named ``fear'') and other, possibly non-heuristic optimization or deliberation procedures for action selection and control. A1--A8 thus also function as demarcation criteria: arbitrarily complex controllers that fail to implement A1--A8 would not qualify as emotion-like. Nevertheless, we may ask whether A1--A8 are still too primitive to capture anything essential about emotion at all. Our human emotional life encompasses far more than these architectural principles: we experience feelings, reflect on them, narrate them to ourselves and others, and integrate them into our sense of self. This richness may lead us to see A1--A8 as too impoverished to deserve the label. But our claim is not that A1--A8 are \emph{sufficient} for the full human experience of emotion---only that they represent a \emph{minimal core} that should be present before we could call any system emotion-like. Human emotional life has much more; we propose that any emotion-like system should have \emph{at least} this. We return to these questions in \S\,\ref{sec:synth_outlook}, where we discuss a potential biological analogy and also discuss how our methodology may remain robust under adaptations of design elements.

\paragraph{The separation question and design constraints.}

Our analysis can also be motivated by the following considerations: if \emo{} can operate through local heuristics (need-based appraisals and similarity-based retrieval), while consciousness apparently requires complex architectural commitments (global integration, higher-order representation, temporal binding), separation of the two notions may be expected. The surmised computational simplicity of \emo{} would suggest the possibility of early evolutionary emergence of biological analogues, potentially requiring only minimal nervous systems. The architectural demands of consciousness might instead explain its rarity and late emergence. This perspective reframes the question from whether emotion and consciousness \emph{can} be separated to charting paths by which artificial systems might traverse the gap between them.

Against this backdrop, we attempt to identify a \emph{conservative}, audit-friendly region of design space which avoids features commonly associated with consciousness. We characterize this region by four design constraints R1--R4 distilled from major consciousness theories (detailed mapping in \S\,\ref{sec:GWTHOT}). These constraints target the \emph{absence} of access-enabling features in prominent accounts. The restrictions carve out a small region that should \emph{plausibly} be non-access-enabling under those theories. Consequently, violating R1--R4 would \emph{not} imply access consciousness under these theories. We call implementations satisfying R1--R4 \emph{plausibly access-excluding} under this multi-family proxy: they inhabit a conservative region designed to remove several prominent access-enabling routes, without claiming that consciousness is thereby impossible.

\begin{enumerate}[leftmargin=1.5em,itemsep=0.25em,label=\textbf{R\arabic*}]

\item \textbf{No global, content-general shared broadcast}:
the system contains no mechanism (workspace, shared scratchpad, hub, or router)
that makes internally generated representations \emph{flexibly reusable} across
heterogeneous subsystems.
\item \textbf{No metarepresentation}: no higher-order representations; in particular, no representations \emph{about} having internal states.
\item \textbf{No autobiographical consolidation} or autobiographical temporal integration across episodes: 
forbid constructing, persisting, exporting, or re-ingesting identity- or time-anchored cross-episode summaries (including as retrieval keys, routing/addressing signals, or shared context variables).
\item \textbf{Bounded learning}: credit assignment remains within module boundaries; no cross-module meta-optimization that induces global coupling.
\end{enumerate}

R1 (no global broadcast) is intended as an architectural constraint: it checks whether \emph{selected internal content} can become widely available to multiple downstream heterogeneous subsystems, while not forbidding ordinary, fixed dataflow. We treat ``heterogeneous subsystems'' as functionally distinct components with separable code paths and explicit interfaces (e.g., perception, memory, planning, language, action selection), ideally enforceable as process/container boundaries. 
 R1 is \emph{not} satisfied by redefining the entire agent as a single monolithic module. We call a buffer or representation \emph{content-general} if it can carry open-ended task content (e.g., arbitrary perceptual/semantic tokens, rich world-model latents, general-purpose scratchpad text, untyped shared context vectors) such that multiple consumers can repurpose the same representational channel for different functional roles. R1 targets mechanisms which (i) host content-general items, (ii) support multi-consumer access (two or more heterogeneous subsystems can read the same item), and (iii) enable flexible reuse (the item can be exploited for diverse downstream functions without a fixed, role-restricted schema). R1 therefore allows \emph{narrow, typed, role-restricted} point-to-point interfaces provided these channels are not used as general-purpose scratchpads and do not permit arbitrary internal content to be routed to multiple heterogeneous subsystems. In particular, fixed fan-out of \emph{exogenous} inputs provided by sensors to multiple components does not by itself constitute a workspace: R1 targets an \emph{internally generated, content-general} state that is written by one subsystem and then flexibly read and repurposed by multiple heterogeneous consumers.

R2 (no metarepresentation) concerns \emph{controller-internal} higher-order states: the architecture contains no representations \emph{about} having internal states that are consumed by control or learning. In particular, including a language or introspection module may produce outward-facing self-ascriptive text (e.g., ``I feel~\ldots'') or telemetry (e.g., ``fear{=}0.8''). In such a case, safely maintaining R2 would require that such outputs be generated via strictly one-way reporting, without being re-ingested as inputs to scoring, memory keys, routing, or learning. Any feedback path from self-report tokens into control would constitute an explicit R2/R3 pressure point.

R3 does not forbid any temporal processing (e.g., filtering), nor does it prevent the mere time-stamped storage of past episodes: it focuses on using such data to synthesize and employ an autobiographical self-description.

R4 (bounded learning) concerns training-like 
architectural parameter updates and cross-module optimization, not the 
accumulation of emotionally tagged episodes---the latter being intrinsic to 
emotion-like control (A1--A8). We therefore distinguish training/learning of architectural parameters or 
neural network weights (the learning constrained by R4) from episodic 
memory formation, which stores affect-tagged experiences for later 
retrieval without parameter modification.

\paragraph{Research questions.}
Note that A1--A8 and R1--R4 derive from different considerations: the architectural components A1--A8 are motivated by biological and psychological research on emotion (see~\S\,\ref{sec:bioA1toA8}), while the risk-reduction constraints R1--R4 emerge from central features of theories of consciousness (see~\S\,\ref{sec:GWTHOT}).
Against this backdrop, we pose three questions about the separability of affect and consciousness:
\begin{enumerate}[leftmargin=1.5em,itemsep=0.25em]
\item[\textbf{Q1}] \textbf{(Existence):} Does there exist a \emph{separation witness}---an implementation of A1--A8 satisfying R1--R4?
\item[\textbf{Q2}] \textbf{(Stability):} Are there design modifications that preserve witness status, or is it structurally fragile?
\item[\textbf{Q3}] \textbf{(Gradualism):} Can we trace graded paths plausibly increasing access risk?
\end{enumerate}

 When addressing Q1, the core 
risk reductions are R1--R3; we include R4 throughout for completeness since 
capability-driven systems often add learning, and its boundaries matter 
for Q2--Q3.

\FloatBarrier

\paragraph{Contributions and roadmap.}
This is a conceptual exploration at the interface of philosophy and computer science, grounded in neuroscience and psychology. We (1) articulate a biologically motivated, hierarchical dual-source reference model of synthetic emotion; and (2) use plausibility arguments to explore separating affect from access consciousness by means of three separation questions Q1--Q3, related to access-related risk reductions R1--R4. Our analysis rests on concrete architectural designs. With these considerations in mind, we offer a plausible witness for Q1; we find pockets of stability (Q2) and explore graded upgrade paths that threaten separation (Q3). After discussing related work in \S\,\ref{sec:precedents}, motivations from consciousness theories in \S\,\ref{sec:GWTHOT}, and biological drives in \S\,\ref{sec:biology}, we outline a general \emo{} architecture and processing loop in \S\,\ref{sec:A1toA8}, made concrete by a toy example for a potential separation witness. We address questions Q1--Q3 in \S\,\ref{sec:Q1toQ3}, while \S\,\ref{sec:synth_outlook} synthesizes findings and discusses the broader methodological template, including its generalizability beyond our specific choices.


\section{Theoretical Grounding}\label{sec2}
Our approach is motivated by four strands of prior work that, taken together, suggest that \emph{emotion-like control} can be engineered and evaluated without making commitments about phenomenology. First, computational architectures integrate affective and motivational signals into action selection (see~\S\,\ref{sec:precedents}). Second, theories of consciousness provide \emph{architectural features} we interpret as \emph{design constraints} to avoid (see~\S\,\ref{sec:GWTHOT}, motivating R1--R4). Third, heuristic perspectives view emotions as resource-bounded control signals that bias decision making (see~\S\,\ref{sec:biology}, motivating A1--A8 and Q1--Q3). Finally, recent proposals for \emph{indicator properties} and governance frameworks suggest how to operationalize \emph{auditable, non-phenomenological} assessments of complex systems~\citep{butlin2025identifying,Raji2020Audit,NIST2023AIRMF}.

\subsection{Computational precedents}\label{sec:precedents}

We draw on multiple computational traditions that model emotions without presupposing phenomenology or full-blown propositional cognition.
Psychological appraisal theories~\citep{smith1993appraisal,lazarus1991emotion,OrtonyCloreCollins1988OCC} hold that emotions arise from evaluating situations along dimensions such as goal-conduciveness, accountability, and coping potential. Computational implementations such as EMA~\citep{MarsellaGratch2009EMA} operationalize these evaluations, driving emotion dynamics and coping responses via propositional machinery (belief tracking, causal attribution, expectation maintenance). The Belief-Desire-Intention (BDI) framework~\citep{RaoGeorgeff1995BDI} provides a practical reasoning architecture in which agents maintain beliefs about the world, desires (goals), and intentions (committed plans). Emotional extensions to BDI-style deliberation, such as EBDI~\citep{jiang2007ebdi} and the WASABI architecture~\citep{BeckerAsano2008WASABI}, introduce affective states drawing on dimensional models such as PAD~\citep{Mehrabian1996PAD}. The LIDA architecture~\citep{Franklin2014LIDA} integrates attention and global-workspace-style access with motivation, emotion, and learning in a biologically inspired systems-level design.

Relative to these precedents, our architecture (i) \emph{eschews} propositional reasoning in favor of \emph{situation-category keys} and \emph{policy hints} as low-bandwidth surrogates for ``what to do next''; (ii) uses affect channels to bias behavioral policy selection directly rather than constructing discrete narratives or modulating a deliberation cycle; and (iii) employs \emph{episodic transfer}---similarity-based retrieval over situation categories that supplies policy-class priors without forward models or multi-step counterfactuals. The resulting pipeline (observations $\rightarrow$ categories $\rightarrow$ policies $\rightarrow$ actions) functions as a computational ``complexity firewall''. This deliberate simplicity reflects our methodological choice of adding complexity only when required (see~\S\,\ref{sec:synth_outlook}). At the same time, A1--A8 remain richer than purely reactive architectures such as Braitenberg vehicles~\citep{braitenberg1984vehicles}. They establish a floor above stimulus-response control while staying below deliberative reasoning.\footnote{Such parsimony would be misguided if emotion-like control inherently demanded richer machinery---but this is not universally assumed. Work on subcortical emotional systems has been interpreted as suggesting that adaptive affective control can arise through ancient, non-propositional mechanisms without cortical elaboration~\citep{panksepp1998affective,panksepp2012archaeology}, providing biological motivation for examining minimal architectures and their functional capabilities.}

\subsection{Design constraints from consciousness theories}
\label{sec:GWTHOT}

We \emph{motivate} the design risk-reduction criteria R1--R4 primarily from two well-developed families of access-oriented accounts---Global Workspace Theory / Global Neuronal Workspace (GWT/GNW) and Higher-Order Thought (HOT) theories---and then note convergent pressure toward similar risk reductions from several additional frameworks. Throughout, R1--R4 are intended as \emph{conservative, audit-friendly risk-reduction criteria}: satisfying them does not establish the absence of consciousness, but plausibly removes multiple architectural routes that these accounts treat as enabling access-like consciousness. Because our operational target is \emph{access-like} consciousness, R1--R4 should be read as risk reductions targeting \emph{access-oriented} routes; they are not intended as exhaustive tests, nor do they address theories that primarily target \emph{phenomenality} (but see the methodological extensions discussed in \S~\ref{sec:synth_outlook}).

\paragraph{Global Workspace Theory (GWT/GNW).}
If GWT/GNW is approximately correct~\citep{DehaeneNaccache2001Workspace,baars2005global,Mashour2020GNW}, access consciousness involves (i) broad availability of selected content via workspace-like broadcast and (ii) stabilization of that content over an operational time window (often discussed in terms of ignition/maintenance dynamics).
 R1 \emph{(no content-general shared global broadcast)} targets the creation of workspaces of selected internal content for multiple heterogeneous consumers, while still allowing narrow, typed, role-restricted control signals flowing through fixed interfaces.
 R3 \emph{(no autobiographical consolidation)} is introduced more cautiously: while workspace theories primarily concern within-window stabilization, engineered systems often extend short-window maintenance into longer-horizon, cross-episode context (e.g., persistent summaries or identity-anchored states that are repeatedly re-ingested). R3 should block this escalation pathway by forbidding autobiographical consolidation. R4 \emph{(bounded learning)} further limits training-driven emergence of broadcast-like routing or implicit workspace structures (e.g., shared latents or routers that effectively recreate multi-consumer availability).

\paragraph{Higher-Order Thought (HOT) theories.}
Under HOT accounts~\citep{Rosenthal2005ConsciousnessMind,lau2011hot}, awareness requires representing one's own mental states (e.g., representing that one is afraid, not merely being afraid). R2 \emph{(no metarepresentation)} should therefore block a central HOT route by forbidding representations \emph{about} having internal states, including internal self-ascriptions that feed back into control or learning. This constraint is further motivated by recent proposals for \emph{conscious} emotional experience. \citet{ledoux2017higher} and \citet{rolls2014emotionbook,rolls2014emotionprecis} argue that conscious emotional \emph{feelings} require higher-order representations.  
 We read both positions as support for treating R2 as a risk-reduction criterion for \emph{conscious} affect. R3 \emph{(no autobiographical consolidation)} provides complementary protection in a practical sense: robust higher-order self-attribution in implemented systems commonly recruits persistent self/identity variables and diachronic bookkeeping (even if not mandated by the core theory). By preventing autobiographical consolidation and re-ingestion of identity/time-anchored summaries, R3 removes a common scaffold by which higher-order access can become temporally extended and self-anchored. R1 \emph{(no content-general shared global broadcast)} offers an additional constraint, since higher-order representations typically require some form of accessibility to their first-order targets across functional components.

\paragraph{Convergent support from other frameworks.}
Several additional approaches, while differing substantially in their explanatory commitments, emphasize architectural moves that also pressure parts of R1--R4:

\begin{itemize}[leftmargin=1.25em,itemsep=0.25em]

\item \textbf{Predictive Processing and Active Inference}~\citep{hohwy2013predictive,seth2013interoceptive,letheby2017self,friston2018deep} often emphasize hierarchical generative models with temporal depth. On many readings, consciousness-relevant capacities involve relatively deep temporal integration and (in some proposals) explicit self-model variables; R3 (limiting autobiographical stitching of cross-episode narratives) and R2 (blocking explicit higher order self-models) target these routes.

\item \textbf{Autonoetic consciousness}~\citep{Tulving1985MemoryConsciousness} concerns mental time travel and re-experiencing past episodes as one's own. This capacity depends constitutively on autobiographical memory and persisting self-continuity---precisely the kind of cross-episode stitching that R3 targets.

\item \textbf{Attention Schema Theory (AST)}~\citep{GrazianoWebb2015AST} proposes that subjective awareness depends on an internal model of attention. If such a schema is an explicit model \emph{about} the system's own attentional processing and is consumed by downstream control, it pressures R2 (metarepresentation), and, in many implementations, also R1 insofar as the schema must be available across functional subsystems.

\end{itemize}

\noindent
Taken together, these different perspectives suggest that R1--R4 track a cluster of architectural moves that are \emph{recurrently flagged} as consciousness-relevant in the literature, rather than being tied to any single theoretical commitment.

\subsection{Biological Grounding}\label{sec:biology}

From a biological standpoint, brains are optimized for adaptive behavior under 
evolutionary pressure, not rationality \textit{per se}. We follow the tradition 
that views emotions as evolutionarily selected heuristics for navigating complex 
environments under constraints of limited time and computational 
capacity~\citep{simon1956rational,simon1967motivational,gigerenzer2001bounded}. 
Stronger emotions exert greater control precedence, overriding deliberation~\citep{frijda1986emotions}, suggesting adaptive benefits for rapid decision-making.

\subsubsection{Biological motivation for A1--A8}\label{sec:bioA1toA8}
Our architectural principles A1--A8 draw on biological, neurological, and psychological evidence to identify possible functional cores of \emo{}. The evidence below is offered as motivational grounding, not derivation: we aim to show that A1--A8 are consistent with plausible readings of diverse findings, while acknowledging that alternative interpretations and architectures remain possible.

\paragraph{Hierarchy and abstraction (A1, A5--A6).} Emotions bias action through ``action readiness'': in social animals, fear biases avoidance, joy biases approach, etc.~\citep{frijda1986emotions}. Such biases trigger broad behavioral modes instantiated into context-appropriate actions~\citep{fanselow1994neural}. Categorical abstraction addresses a prerequisite for generalization since raw sensory states rarely repeat exactly.
The requirement that emotional processing produce an intervening state that 
enables flexible instrumental action distinguishes emotion from reflexes and fixed action 
patterns~\citep{rolls2014emotionbook,rolls2014emotionprecis}.

\paragraph{Dual mechanisms (A2--A4).} Subcortical circuits generate immediate 
responses from current discrepancies: homeostatic needs~\citep{damasio1994descartes}, 
safety violations~\citep{ledoux2012rethinking}, and social 
affective needs~\citep{panksepp1998affective} (motivating A2).  Retrieved 
episodes bias current decisions through similarity-based reactivation: transference research demonstrates that affect associated with past significant others transfers to similar individuals~\citep{Andersen1994Transference}, while behavioral evidence shows past choices bias decisions even without explicit recollection~\citep{bornstein2017reminders}. Our architecture generalizes these findings to situational similarity (motivating A3). The existence of immediate 
and memory-based responses suggests dual-source integration\footnote{This design remains compatible with cases like patient K.C., who lost all conscious access to past episodes yet retained other cognitive functions~\citep{Tulving2002EpisodicMemory}: (a) our episodic retrieval may fail and affect can then derive solely from current needs (graceful degradation), and (b) our episodes influence behavior exclusively through transferred affect and behavioral biases, not surfacing as recollections or basis for deliberation.} (motivating A4).

\paragraph{Reappraisal and storage (A7--A8).} Post-action reappraisal generates feedback updating future responses. The somatic marker hypothesis can be read as partial support for emotional outcomes of past decisions influencing current choice~\citep{damasio1994descartes,bechara2000decision}, motivating the use of affective tags in our episodic memory. Emotionally salient experiences are preferentially consolidated~\citep{mcgaugh2004making,kensinger2004remembering}; notably, affective vividness outlasts factual accuracy in ``flashbulb memories''~\citep{talarico2007flashbulb}. Affective tags may thus bias future behavior even as episodic detail fades~\citep{rolls2014emotionbook,rolls2014emotionprecis} (motivating A7--A8).

\subsubsection{Biological motivation for Q1--Q3}\label{sec:bioQ1toQ3}

Multiple lines of evidence suggest partial independence between functional and phenomenal affect. Subliminal priming demonstrates valence effects below awareness~\citep{wiens2006unconscious}. Affective blindsight shows emotional responses without conscious perception~\citep{CeleghinDeGelderTamietto2015}. Liking/wanting dissociations reveal unconscious hedonic 
reactions~\citep{berridge2003unconscious}. \citet{rolls2014emotionbook,rolls2014emotionprecis} argues that action selection may proceed through implicit routes without requiring conscious 
feelings. \citet{ledoux1996emotional} suggested a fast/slow distinction, motivating us to consider whether affect may act independently of detailed understanding.

Some neurological cases appear to suggest dissociability~\citep{bechara2000decision,feinstein2013co2}. 
Given ongoing debates about the relationship between emotion and consciousness---including subcortical affect systems~\citep{panksepp1998affective,panksepp2012archaeology}, integration with predictive processing~\citep{seth2013interoceptive}, proposed dissociations~\citep{TamiettoDeGelder2010}, and constructionist accounts~\citep{barrett2017construction}---we treat Q1--Q3 as open empirical questions rather than presupposing any particular resolution.

Multiple lines of research indicate emotional memories undergo offline processing~\citep{walker2009role,mcgaugh2004making,revonsuo2000dreams}, motivating our reconciliation process (further motivating Q2 and architectural modification M1 in \S\,\ref{sec:Q2}), though we claim no functional equivalence to biological sleep or dreaming.

The progression from simple state-modulation in \textit{C.\ elegans} through episodic memory in mammals~\citep{ginsburg2019evolution,adolphs2018structure} motivates Q3: systems might progressively approach consciousness-relevant features when transitioning from reactive to more sophisticated forms of predictive control. This progression also suggests a background hypothesis underlying Q1--Q3 and developed further in \S\,\ref{sec:synth_outlook}: emotion-like control may constitute an intermediate computational layer that can operate without access consciousness (Q1--Q2; \S\S\,\ref{sec:Q1}--\ref{sec:Q2}), yet becomes intertwined with consciousness in certain systems (Q3; \S\,\ref{sec:Q3}).

If \emo{} appears to operate through relatively simple mechanisms in biology, perhaps artificial systems can achieve similar functionality while avoiding consciousness-associated features. This motivates constructing an architecture that instantiates hierarchical dual-source principles (A1--A8) while respecting constraints R1--R4.

\section{An Architecture for \EMO{}}\label{sec:A1toA8}

We now translate the biological principles and definitional components into a concrete, example architecture.

\subsection{A primitive emotion-like control implementation}
\label{sec:primitiveexample}

The following end-to-end controller implements key design moves that follow  A1--A8. The system hierarchically groups situations into \emph{categories} which relate to \emph{policy hints} (behavioral modes) that get instantiated as concrete actions selected using emotional biases.\footnote{A detailed discussion of the controller is provided in SI, Part I.}

\paragraph{Setting.}
The algorithm in Fig.~\ref{fig:control_flow} is applicable in a wide range of contexts. For ease of understanding, the following discussion may be visualized in the context of a simplified concrete example:  consider an agent in a 2-D crowd scene. At each step $t$ it receives a compact observation vector $\myvect{x}_t\in\mathbb{R}^{d}$ (e.g., coarse bearings/distances to peers, proprioception, current heading, location, speed), and selects a fine-grained action $u\in\mathcal{U}$ (e.g., move in heading $\theta$ or pause). Internal affective variables are used only as control signals. We may group affect attributed to our agent into low-dimensional affect channels  like valence $\bm{v} \in [-1,1]^b$, corresponding emotional intensity/magnitude $\bm{m}\in [0,1]^b$, arousal $a\in [0,1]$, and motivation/drive $\bm{d}\in [-1,1]^\ell$:
\begin{equation}
\myvect{z}_t :=   [\bm{v}_t, \bm{m}_t, a_t, \bm{d}_t], 
\end{equation}
where the bracket denotes concatenation. For our actor we may imagine two affect channels $b = 2$ with emotional valence $\bm{v}=(v_1,v_2)$ and magnitude $\bm{m}=(m_1,m_2)$, where the pair $(v_1,m_1)$ encodes lonely-like emotions, and $(v_2,m_2)$ stands for crowded-like emotions. The $\ell$ drives $d_1,\ldots,d_\ell$ provide a measure of motivation and correspond in number to $\ell$ different need channels. In our simplified example, $n_{1,t}$ may represent a need for affiliation or bonding, and $n_{2,t}$ may represent a need for independence; both may depend on the observed geometric peer layout.\footnote{For other examples, see SI, Parts I, I.A-B.}

\begin{figure}[!t]
\centering
\scalebox{0.7}{%
\begin{minipage}{\textwidth}
\begin{align}
&\textbf{// A1: Classify Situations} \nonumber\\
1.~&\myvect{x}_t \leftarrow \text{observe}(\mathscr{W}_t) \quad && \text{(raw sensory state)} \nonumber\\
2.~&(\myvect{c}_t, \myvect{y}_t) \leftarrow \text{categorize}(\myvect{x}_t)  && \text{(category and parameters)} \nonumber\\
\nonumber\\
&\textbf{// A2: Need Appraisal} \nonumber\\
3.~&\myvect{n}_t \leftarrow \text{assess\_needs}(\myvect{c}_t, \myvect{y}_t; \myvect{x}_t ) && \text{(situational needs)} \nonumber\\
4.~&\myvect{z}^{\text{need}}_t \leftarrow [\myvect{v}^{\text{n}}_t, \myvect{m}^{\text{n}}_t, a^{\text{n}}_t, \myvect{d}^{\text{n}}_t] = f_{\text{affect}}(\myvect{n}_t) && \text{(affect from needs)} \nonumber \\
5.~&\myvect{h}^{\text{need}}_t \leftarrow \text{hints}^{\text{(need)}}(\myvect{z}^{\text{need}}_t) \qquad (\text{e.g.} \, h^{\text{need}}_t(\pi)=\big[\mymat{H}_{\text{n}}  \, \myvect{d}^{\text{n}}_t \big]_\pi \,)\quad &&\text{(policy hints from needs)} \nonumber\\
\nonumber\\
&\textbf{// A3: Episodic Retrieval} \nonumber\\
6.~&\mathcal{N}_K(t) \leftarrow \text{top-}K\{\myvect{c}_j : \myvect{c}_t^T\myvect{c}_j\} ~~\text{from memory}~\mathcal{M}~~&& \text{(similar episodes)} \nonumber\\
7.~&\myvect{z}^{\text{mem}}_t \leftarrow \sum_{j \in \mathcal{N}_K(t)} w_j \myvect{z}^{\text{tag}}_j \quad \text{with} \quad \myvect{z}^{\text{tag}}_j\leftarrow\text{combine}(\myvect{z}_j,\myvect{z}^\ast_j) &&\text{(episodic  affect)} \nonumber\\
8.~&h^{\mathrm{mem}}_t(\pi)
\leftarrow \sum_{j\in\mathcal{N}_K(t)} w_j\,\text{succ}^\ast_j\,h_j(\pi),  ~~(w_j\text{ sum to }1)
  && \text{(episodic policy hint)} \nonumber\\
\nonumber\\
&\textbf{// A4: Affective Integration} \nonumber\\
9.~&\myvect{z}_t \leftarrow F(\myvect{z}^{\text{need}}_t, \myvect{z}^{\text{mem}}_t)  && \text{(fused affect)} \nonumber\\
10.~&\myvect{h}^{\text{aff}}_t \leftarrow \text{hints}^{\text{(affect)}}(\myvect{z}_t) \qquad (\text{e.g.} \, h_t^{\text{aff}}(\pi)=\big[ \mymat{H}  \, \myvect{z}_t \big]_\pi  \,)\quad && \text{(policy hints from affect)} \nonumber\\
11.~&\myvect{h}_t \leftarrow \alpha_n\myvect{h}^{\text{need}}_t + \alpha_m\myvect{h}^{\text{mem}}_t + \alpha_a \myvect{h}^{\text{aff}}_t && \text{(fused policy hints)} \nonumber\\
12.~&{q}_t(\pi) \leftarrow \text{softmax}_\pi( \myvect{h}_t/ \tau_1(a_t)) \quad && \text{(policy probabilities)} \nonumber\\
\nonumber\\
&\textbf{// A5: Policy Instantiation} \nonumber\\
13.~&{\tilde s}_\pi(u) \leftarrow \text{policy\_to\_action}_\pi(u ; \myvect{c}_t, \myvect{y}_t; \myvect{x}_t)  \quad \forall u \in \mathcal{U} \quad && \text{(score actions by policy)} \nonumber\\
14.~&s_t(u) \leftarrow \sum_{\pi} q_t(\pi) \cdot {\tilde s}_\pi(u) \quad \forall u \in \mathcal{U} \quad && \text{(policy weighted scores)}\nonumber\\
\nonumber\\
&\textbf{// A6--A7: Execute and Reappraise} \nonumber\\
15.~&u_t \sim \text{softmax}_u( \myvect{s}_t / \tau_2(a_t)) \quad \text{or } u_t = \arg\max_u s_t(u) && \text{(sample action)}\nonumber\\
16.~&(\myvect{c}^\ast_t, \myvect{y}^\ast_t; \myvect{x}^\ast_t) \leftarrow \text{execute\_observe\_categorize}(\mathscr{W}_t; u_t) && \text{(post-act state)}\nonumber\\
17.~&\myvect{n}^\ast_t \leftarrow \text{assess\_needs}(~\myvect{c}^\ast_t, \myvect{y}^\ast_t; \myvect{x}^\ast_t~) && \text{(post-act needs)} \nonumber\\
18.~& (\myvect{z}^\ast_t,\text{succ}^\ast_t) \leftarrow \text{reappraise}(~\myvect{z}_t; \myvect{x}_t, \myvect{n}_t, \myvect{x}^\ast_t, \myvect{n}^\ast_t~) && \text{(post-act affect$+$success)}\nonumber\\
\nonumber\\
&\textbf{// A8: Episode Storage} \nonumber\\
19.~&e_t \leftarrow ( \myvect{c}_t, \myvect{z}_t, \myvect{h}_t, \myvect{z}^\ast_t, \text{succ}^\ast_t )  && \text{(aggregate episode)} \quad  \quad \nonumber\\
20.~&\mathcal{M} \leftarrow \mathcal{M} \cup \{e_t\}  && \text{(memorize episode)}\nonumber\\
\nonumber 
\end{align}
\end{minipage}%
}
\caption{$\mathcal{S}$: a single iteration (A1--A8) in a hierarchical, dual-source emotion-like control loop for action selection. For a detailed discussion see~Supplementary Information, Part I.}
\label{fig:control_flow}
\end{figure}

\paragraph{Architectural Overview.}
\label{sec:architectural_overview}

The architecture of Fig.~\ref{fig:control_flow} operates across two levels of description---a \emph{concrete level} interfacing with the physical world, and an \emph{abstract level} where emotional signals guide behavioral policy selection---with systematic information flow between them. Information flows ``upward'' through categorical abstraction, where dual-source emotional guidance shapes policy selection at the abstract level, then ``downward'' through policy instantiation into concrete action. Outcomes feed back through reappraisal into episodic storage.

\paragraph{A1:~Upward abstraction: From sensation to situational encoding.}
Raw sensory input $\myvect{x}_t$ rarely repeats exactly across encounters, rendering direct episodic matching brittle. The system therefore first maps the concrete situation into a more abstract \emph{situation-category} $\myvect{c}_t$ that captures relevant structure (e.g., ``moderately crowded,'' ``threat approaching'') while also extracting situation-specific parameters $\myvect{y}_t$ (e.g., direction to nearest crowd, threat bearing) for later use in action instantiation.

\paragraph{A2--A3:~Dual-source emotional hints at the level of abstract policies.}
With the situation categorized, the system consults two sources of affective signals: need-based affect $\myvect{z}^{\text{need}}_t$ and memory-based affect $\myvect{z}^{\text{mem}}_t$.

\emph{Source 1:} Need appraisal. Current homeostatic discrepancies---deviations between actual and target states for affiliation, safety, energy, and similar needs---generate drives carrying both direction and intensity, from which emotional signals (valence, arousal) and policy preferences emerge. In step~3 of Fig.~\ref{fig:control_flow}, need levels are computed from the \emph{current} world situation, including categorized abstractions of observations (e.g., threat levels) and internal control variables (e.g., maintenance levels, physical or chemical sensor readings). Need levels, especially when compared to target values, generate corresponding affect (step~4 of Fig.~\ref{fig:control_flow}). This affect also provides policy hints $h_t^{\text{need}}(\pi)$, which quantify how strongly need-based affect $\myvect{z}^{\text{need}}_t$ favors each behavioral policy $\pi$---broad behavioral modes that admit different concrete instantiations as actions $u$. In our simplified example, such policies $\pi$ may include \textsc{Seek}, \textsc{Avoid}, \textsc{Explore}, \textsc{Flee}, or \textsc{Rest}, while affect components $\myvect{z}^{\text{need}}_t$ react to motivational drives reflecting how far energy, safety, bonding, or independence levels lie from their targets.

\emph{Source 2:} Episodic memory. The situation-category $\myvect{c}_t$ serves as a retrieval key into stored episodes. Memory returns the affective ``coloring'' of similar past situations, $\myvect{z}^{\text{mem}}_t$, along with \emph{policy hints} $\myvect{h}^{\text{mem}}_t$---success-weighted suggestions about which behavioral strategies proved effective under comparable circumstances ($\myvect{c}_t \approx \myvect{c}_j$). When computing episodic affect and policy hints, additional weights $w_j$ may be used to fine-tune the influence of past episodes by their degree of situational similarity. For simplicity, steps 6--8 are depicted as part of the controller $\mathcal{S}$ in Fig.~\ref{fig:control_flow}, even though they would conceptually be performed inside a memory module, which only returns the final affect $\myvect{z}^{\text{mem}}_t$, along with policy hints $\myvect{h}^{\text{mem}}_t$, i.e.~
\begin{equation}
(\myvect{z}^{\mathrm{mem}}_t,\, \myvect{h}^{\mathrm{mem}}_t)
\;\leftarrow\;
\mathrm{Retrieve}_{\mathcal{M}}(\myvect{c}_t),
\label{eq:mem-interface}
\end{equation}

\noindent
where $\mathrm{Retrieve}_{\mathcal{M}}$ is a bounded-output procedure (e.g., top-$K$ similarity search with fixed-schema aggregation) that exposes only $(\myvect{z}^{\mathrm{mem}}_t,\myvect{h}^{\mathrm{mem}}_t)$ to the controller $\mathcal{S}$, possibly with an overall reliability indicator (e.g., degree of similarity to most similar episode).

\paragraph{A4:~Policy selection through affective integration.}
Affective signals from both sources (needs and memory) are fused into a combined affective state $\myvect{z}_t$, which provides hints for behavioral policies. All hint sources (fused affect, needs, memory) contribute to aggregated hints $h_t(\pi)$ for each abstract policy $\pi$, yielding policy preferences $q_t(\pi)$. These preferences $q_t(\pi)$ reflect a continuous-value vote for each policy. Arousal $a_t$ may modulate decisiveness: high arousal sharpens the vote distribution toward commitment, while low arousal encourages exploration.

\paragraph{A5:~Downward instantiation: From policy to action.}
Abstract policies do not specify concrete actions. The system uses situation-specific parameters $\myvect{y}_t$ to \emph{instantiate} policies into action scores $s(u)$. For example, if the policy $\pi = \textsc{Avoid}$ is tentatively considered and parameters $\myvect{y}_t$ indicate that a threat lies to the left, concrete rightward movements $u$ receive high scores $s(u)$. 
This context-dependent mapping ($\myvect{y}_t$) from broad policies $\pi$ to concrete actions $u$ ensures that memory stores policies---general behavioral modes rather than motor sequences---enabling behavioral generalization (e.g., fight or flight) across wide situational variations (different threat positions). This simplified algorithm can thus learn \textit{when} to activate policies, but policy-to-action mappings are assumed innate. Presently, a policy $\pi = \textsc{Flee}$ directs the agent toward a \textit{predetermined}, presumably safer region, analogous to an innate taxis-like response, e.g., movement away from threats.\footnote{See also the harm-avoidance example discussed in SI, Part I. Learning movement directions would require memory-informed policy instantiation, which we leave for future work, see also SI, Part I.B.}

\paragraph{A6--A8:~Execution, reappraisal, and storage.}
The agent executes the action $u$ with the highest score $s(u)$---or sampled  
from the score distribution---observes the resulting state $\myvect{x}^\ast_t$, 
categorizes it ($\myvect{c}^\ast_t$, $\myvect{y}^\ast_t$), and reappraises outcomes 
against needs, producing post-action affect $\myvect{z}^\ast_t$. The 
complete episode $e_t$ comprises the situation-category $\myvect{c}_t$ (which 
also serves as future retrieval key), pre-action affect $\myvect{z}_t$, policy 
hints $\myvect{h}_t$, outcome affect $\myvect{z}^\ast_t$, and success estimate 
$\text{succ}^\ast_t \in [-1,+1]$, where negative values indicate failure. 
Success may be estimated from ``objective'' measures of situational improvement, from need improvements (e.g., drive reduction),   
or from hedonic changes by comparing $\myvect{z}^\ast_t$ to $\myvect{z}_t$. Alternatively, one may compare only the pre- and post-action need-based affect (e.g., if relief from memory-induced fear is considered an artifact). Note 
that the minimal implementation in Fig.~\ref{fig:control_flow} stores the fused 
hints $\myvect{h}_t$; but memory could additionally store the selected policy 
$\pi_t$ for each episode and use this information in the weighted averaging of Steps~7--8 for 
stricter credit assignment (see~SI, Part~I.B for details). 
Thus, every encountered situation is recorded in 
condensed, categorized form ($\myvect{c}_t$) together with affective tags, policy hints, and success 
measures, ready for use in future policy and action selection.\footnote{Our example may maintain memory-internal housekeeping metadata (e.g., an age counter or write index) for pruning or deduplication. Such metadata is not part of the episode fields returned to the controller $\mathcal{S}$.}

The architecture can operate standalone or augment existing systems. Hybrid controllers could fuse affect-based action scores $s_{\text{emo}}(u)$ with scores $s_{\text{delib}}(u)$ from deliberative planners, using arousal or affect to gate exploration, persistence, or risk sensitivity. Alternatively, the low-cost $s_{\text{emo}}(u)$ could serve as initial suggestions for deliberative planners (mirroring volitional-like control). The affective path aims at supplying temporal continuity (through continuity in arousal and affect), coherence across competing needs via information fusion, and contextual adaptation.

\section{Examining the Separation}\label{sec:Q1toQ3}

We now return to the three questions posed in the introduction, examining whether and how \emo{} can be implemented while avoiding consciousness-enabling mechanisms. This section demonstrates that the architecture from \S\,\ref{sec:A1toA8} can serve as a separation witness (Q1), identifies stable modifications preserving this status (Q2), and explores gradual paths toward consciousness-relevant thresholds (Q3).

\subsection{Q1: The Architecture as Separation Witness}\label{sec:Q1}

Can emotion-like control following design principles A1--A8 be implemented to satisfy R1--R4? 
We argue that it can, and establish a separation witness starting from the primitive example outlined in Fig.~\ref{fig:control_flow}, further restricting attention to \emph{minimal} implementations that respect the following conservative safety envelope:

\paragraph{Safe Interface Contract (SIC).}
We formalize interface constraints on memory $\mathcal{M}$ and controller $\mathcal{S}$ in Fig.~\ref{fig:control_flow}:
\begin{enumerate}[leftmargin=5em,itemsep=0.15em]
\item[\textbf{SIC-1}] \textbf{Single-reader access (memory cannot serve as a workspace).}
Any episode in $\mathcal{M}$ (including caches) has exactly one memory-external heterogeneous reader: $\mathcal{S}$.
This is a sufficient condition preventing $\mathcal{M}$ from functioning as a content-general shared workspace, and preventing the controller $\mathcal{S}$ from benefiting from indirect access to external systems using the same memory (no workspace, R1).
\item[\textbf{SIC-2}] \textbf{Key hygiene (step-local).} Retrieval keys $\myvect{\kappa}_t=\myvect{c}_t(\myvect{x}_t)$ depend only on current
observation content (no identity/time/affect fields; see~R1--R3).
\item[\textbf{SIC-3}] \textbf{No cross-episode summaries to the control-loop.} Offline maintenance inside $\mathcal{M}$ creates no autobiographical stitching of cross-episode aggregates in any buffer read by controller $\mathcal{S}$ (R3).
\item[\textbf{SIC-4}] \textbf{Gradient/optimizer isolation.} Frozen deployment. For any pre-deployment training: no loss outside $\mathcal{S}$ depends on episodes or derivatives thereof; optimizers are disjoint; telemetry containing previous episode data is never used to train modules (R4).
\item[\textbf{SIC-5}] \textbf{Memory access discipline.} \emph{Write only} at A8 (step~20) and \emph{read only} at A3 (step~6).
The observation buffer $\myvect{x}_t$ at A1 is exogenous (world/interoception) and \emph{excludes} introspective reads of
$\mathcal{M}_t$. Retrieval at A3 is single-pass with bounded outputs (R1, R3).
\end{enumerate}

\paragraph{Additional deployment assumptions for the witness.}
Figure~\ref{fig:control_flow} specifies the control flow (A1--A8) but does not \emph{by itself} enforce the
Safe Interface Contract (SIC). The core principles of \emo{} are compatible with a wide range of design choices regarding interfaces and  intertwining with other processes. Because Q1 is an \emph{existence} claim, it suffices to show it for an SIC-compliant implementation, provided such implementations exist.

\medskip
\noindent
\textbf{R1--R4 compatibility:} 
Under SIC-compliant implementations, the controller in Fig.~\ref{fig:control_flow} satisfies R1--R4 and becomes a separation witness for question Q1.

{\small
\emph{Sketch of Argument:}
R1: SIC-1/2/5 enforce a single, bounded read path from memory, preventing workspace/broadcast.
R2: Internal affect variables are numeric control signals used only inside scoring/fusion maps in $\mathcal{S}$; the witness contains no module that re-encodes them as higher-order propositions (self-ascriptions). In addition, SIC-4 assumes frozen deployment (no deployment/post-deployment training), and  also prevents episode-derived telemetry from becoming a training signal for modules.
R3: SIC-2/3 keep keys step-local and prevent write-back of cross-episode aggregates; SIC-5 blocks introspective read loops that could stitch episodes into an autobiographical state.
R4: SIC-4 enforces optimizer isolation and prevents cross-module learning via episode data leakage; SIC-5 further constrains read/write pathways that could otherwise act as unintended training interfaces.
}

\medskip
\noindent
\textbf{Existence of SIC compliance:} 
The argument above covers implementations satisfying the Safe Interface Contract (SIC). SI, Part~I complements this argument with a concrete worked instance (keys, hints, policy-to-action templates), providing an explicit constructive toy model for SIC-compliant implementations and thus completing our affirmative answer to \textbf{Q1} at the single-agent level. The toy model corresponds to ``hardwiring'' or ``freezing'' a pre-specified controller (no training, modeling innate systems) and ensuring it respects a strict read/write discipline. The constructed witness is not only architecturally and parametrically frozen but also deliberately primitive regarding categorical abstraction (implementable even via discrete binning), need-to-affect mapping (implementable as a linear transform), and episodic guidance and information integration (implementable via bounded $k$-NN retrieval, weighted sums, and softmax selection). 
To be clear, our worked instance is a \emph{hardwired} and \emph{standalone} $\mathcal{S}$+$\mathcal{M}$ controller with the SIC enforced without integrating language or deliberative planning, or other heterogeneous subsystems.

\medskip

Given that we can construct such primitive witnesses, it may be worth re-emphasizing that we should not conflate \emph{sophistication} with \emph{membership in the target architectural class}. The point of the Q1 witness is precisely that an implementation can satisfy A1--A8 while remaining compatible with R1--R4. Whether this primitiveness may even hint at corresponding biological hypotheses is discussed in \S\,\ref{sec:synth_outlook}.

\paragraph{Levels of description.}
A feature of functionalist accounts is that architectural features of access consciousness might emerge at higher organizational levels through agent interactions. Block's ``China Brain'' thought experiment illustrates this~\citep{block1978troubles}: could a population simulating neurons with brain-like connectivity give rise to collective consciousness? In our setting, while the witness construction demonstrates R1--R4 compliance for a \emph{single} agent, we may ask whether networks of artificial emotion-like ``neurons''---each operating via A1--A8 and satisfying R1--R4---might instantiate integration at the network level. Crucially, our methodology can be repeated at each hierarchical level: checking A1--A8 for emotion-like control and auditing R1--R4 for access consciousness.

\subsection{Q2: Stable Design Modifications (Plausibility by Examples)}
\label{sec:Q2}

\noindent
Having established the Q1 separation witness under strict R1--R4, we argue for \emph{stability} by concrete, auditable modification examples M1--M3.\footnote{SI, Part II provides complete details for M1, as well as further modification examples M4--M5.}
Each modification below alters functionality without departing from the conservative region defined by R1--R4. Where relevant we give (i) a minimal mathematical specification, (ii) an explicit note on changes compared to Fig.~\ref{fig:control_flow}, and (iii) \emph{audit hooks} that implementers can run to verify compliance. None of the examples presupposes a global workspace, higher-order self-representations, or cross-episode autobiographical state.

\vspace{0.4em}
\paragraph{M1: Offline periodic memory reconciliation (``dream''/``sleep'').}
\emph{Design.} Run a periodic \emph{memory-internal} maintenance pass that leaves the controller $\mathcal{S}$ and its interfaces unchanged. The controller still performs the same bounded read at A3 and write at A8, and memory returns the same fixed-schema fields; the maintenance pass may only prune episodes and/or update \emph{per-episode} fields (e.g., $\successtag$) within that schema. Intuitively, memory keeps the most useful examples and makes success tags reflect medium-term consequences, while the controller and its read/write discipline remain unchanged.
The maintenance pass:\\
    \emph{(i)}  deduplicates near-identical episodes and prunes similar pre-action episodes while \emph{keeping high-success exemplars}, optionally per policy,\\ 
     \emph{(ii)} \emph{prunes} episodes by age or retrieval frequency, with success/emotional intensity based exceptions (``flash-bulb'' exceptions); and\\
     \emph{(iii)}  \emph{re-estimates} single-step past success tags using their medium/long-term horizon outcomes through a bounded, future-informed aggregate (short-term gains/losses versus long-term losses/gains). Distribute a small backward bonus for rare, very-high-success events to preceding actions. 

\emph{Loop impact.} None: the read/write discipline of SIC-5 (read only at A3, write only at A8) and the frozen-deployment assumption of SIC-4 remain satisfied.

\emph{Audit hooks.}  (R1) no broadcast, (R2) no metarepresentations (R3) no autobiographical stitching; only per-episode scalar rewrites; any success edits remain atomic per episode; no new vectors or keys are created, and retrieval keys remain step-local, and (R4) no re-training, only memory module-local updates. This example should highlight that memory-internal maintenance and success re-evaluation, even if concerning multiple episodes, would not by itself imply violations of any of R1--R4.

\vspace{0.4em}
\paragraph{M2: ``Mood''-like temporal smoothing (local integrator).}

\emph{Design.} A leaky buffer biases $\mathcal{S}$ by storing a time-averaged emotion which is neither persisted as an episode nor used as a retrieval key:
\begin{align}
\bar{\myvect{z}}_t \;=\; \gamma\,\bar{\myvect{z}}_{t-1} \;+\; (1-\gamma)\,\myvect{z}_t, 
\qquad \gamma\in[0.9,0.99].
\end{align}

\emph{Loop impact.} Affective integration A4 may use $\bar{\myvect{z}}_{t-1}$ as a controller-internal input to fusion/scoring; retrieval keys remain step-local (SIC-2), and the memory read/write discipline remains unchanged (SIC-5).

\emph{Audit hooks.}  (R1--R2) No additional heterogeneous consumers are introduced, and $\bar{\myvect{z}}$ is not re-encoded into higher-order reportable state tokens. (R3)  While the leaky smoothing of affect implies long time-window averages, $\bar{\myvect{z}}$ is never aggregated into an autobiographical consolidation. 
In particular, (i) it is strictly controller-local, (ii) it is not used as retrieval key or situation-category, (iii) it is not stored inside episodes, and (iv) it has an explicitly bounded time constant (auditable as an effective temporal receptive field). (R4) For innate controllers or frozen deployment (SIC-4), no training occurs and the leaky integrator is simply a fixed runtime mechanism. If pre-deployment training does occur and incorporates $\bar{\myvect{z}}$, gradients must remain module-local and must not backpropagate through the temporal buffer to couple earlier timesteps with later module updates.
This example should illustrate that simple temporal filtering does not by itself imply consolidation into an autobiographical narrative.

\vspace{0.4em}
\paragraph{M3: Trait-like constant modulators.}
\emph{Design.} Introduce per-actor parameters $\Omega\in\mathbb{R}^f$ that \emph{parametrize}
only local maps inside $\mathcal{S}$, without changing interfaces or keys. For example, we may modify the need-based affect-to-hint-mapping 
\[
\myvect{h}^{\text{need}}_t \leftarrow \text{hints}^{\text{(need)}}(\Omega,\,\myvect{z}^{\text{need}}_t).
\]
The values of $\Omega$ may be \emph{interpretable to observers} as ``trait-like''
(for alignment/telemetry purposes), but internally they are just \emph{control gains}.

\emph{Runtime immutability.} During deployment, $\Omega$ remains \emph{constant}, consistent with SIC-4 (frozen deployment / optimizer isolation).

\emph{Loop impact.} None on A1--A8 interfaces: the Safe Interface Contract is unchanged, including SIC-5 (read only at A3, write only at A8). The only effect of $\Omega$ is via the fixed parametrization of bounded, controller-internal maps; retrieval keys and
episodes are unchanged.

\emph{Audit hooks.} 
(R1: no content-general workspace) No content-general shared scratchpad/hub/router is introduced: $\Omega$ only parametrizes fixed, controller-internal maps.
Any shared signals remain narrow, typed control variables (gains, policy parameters, fixed-schema scores), and there is no global attention, message bus, or routing mechanism that would expose a content-general state to multiple heterogeneous subsystems.
(R2: no metarepresentations) Static graph check: no buffer read by $\mathcal{S}$ encodes any token whose semantics is ``my $\Omega$ is \dots''. Episodes continue to store only  situation classes, affect
$\myvect{z}$, policies, and success tags to support similarity-based control. Even though this can lead to non-intuitive 
behavior (for example to \emph{counteract} ``disadvantageous'' parameterizations $\Omega$ because ``non-social''
settings may underperform during deployment and memory-based affect may eventually bias  toward \textsc{Seek}), any such shifts result
from \emph{world-facing} evidence and remain first-order: no variable \emph{about}
$\Omega$ is constructed, stored, or consumed by the action-selection loop (changes in behavior would result without explicit or implicit situational meta-analyses). 
(R3: no temporal integration into cross-episode summaries) $\Omega$ is never written to episodes, never used in keys, not time-indexed. Keys remain step-local functions of $\myvect{x}_t$ (and permitted step-local transforms).
(R4: bounded learning): No optimizer for $\Omega$ during deployment.
This example should illustrate that parameterizations might coherently modulate ``behavior'', yet remain first-order without introducing a higher order self-model.

\subsection{Q3: Gradual Paths Toward Consciousness Risk}\label{sec:Q3}
\paragraph{The gradualism question.}
Having delineated a conservative region in which R1--R4 remain intact, making access consciousness unlikely under our proxy, we now ask whether one can trace paths that gradually increase access-consciousness risk. While we cannot demonstrate consciousness emergence (there is no universally accepted operational test), we can argue that our framing through risk-reduction criteria R1--R4 allows for a natural answer to question Q3: the combined risk-reduction criteria R1--R4 represent the strictest version by establishing many simultaneous constraints. As constraint violations accumulate, risk plausibly increases. Though the precise mapping remains uncertain, we can establish a partial hierarchical ordering of different design paths: any proper subset of constraints $\{\text{R}_1,\ldots,\text{R}_m\} \subsetneq \{\text{R}_1,\ldots,\text{R}_n\}$ requires fewer constraints ($m<n$) to be simultaneously met, and would thus plausibly be \emph{less strongly} access-avoiding. Each abandoned risk-reduction criterion amounts to permitting specific capability upgrades. Hence, risk-reduction criteria may indicate which design elements we may focus on when asking whether—and which—design moves may \emph{gradually} shift toward access consciousness. They can form graded ``drift axes'' in design space which allow us to chart design \emph{trajectories} that leave the conservative region, plausibly in the direction of increased access-consciousness risk.

\subsubsection{Self-modeling as a graded path}\label{sec:q3-self}
A prominent trajectory concerns \emph{self-modeling} \citep[see][]{gerrans2024avatar}. We distinguish a ladder of design moves that gradually pressure R2 (and eventually R1/R3):

\begin{enumerate}[label=\textbf{S\arabic*}, leftmargin=1.6em, itemsep=0.3em]
\item \textbf{Self-free (task-facing only).} The system tracks exogenous/task variables; affect is first-order control (needs/drives $\rightarrow$ policies). Compatible with R1--R4 (our Q1 witness baseline).
\item \textbf{Proto-self (body schema/calibration).} Local proprioceptive calibration signals \emph{inside} $\mathcal{S}$ to stabilize control; not stored or keyed; not reportable. Still first-order; typically R1--R4 compatible if kept private and myopic.
\item \textbf{Episode-linked identity.} Persistent identifiers (or profile features) linking episodes over time. Begins to pressure R3 (temporal autobiographical consolidation). If such identifiers enter $\mathcal{S}$ inputs or keys, R2/R3 risk rises.
\item \textbf{Narrative self-model.} Cross-episode summaries that \emph{represent} the agent’s own states, capabilities, or history; often coupled to broadcast. This directly risks R2 (self-ascriptive tokens), R3 (autobiographical stitching), and frequently R1 (workspace-like hubs).
\end{enumerate}

\noindent
\emph{Takeaway.} The step from \textbf{S2} to \textbf{S3}/\textbf{S4} aligns with concerns based on predictive processing accounts: moving from local calibration to \emph{explicit} self-tokens and narrative integration. Our witness design remains at \textbf{S1} (optionally \textbf{S2}) by construction.

\subsubsection{Separate gradual paths}\label{sec:q3-orthogonal}

\paragraph{Integration pressure (broadcast).}
Demands for cross-modal coordination and routing can gradually erode R1:
\begin{itemize}[leftmargin=1.2em,itemsep=0.2em]
\item \textbf{B1:} Separate processors, narrow typed interfaces.
\item \textbf{B2:} Shared encoders at interfaces (risk: stealth hubs).
\item \textbf{B3:} Common latent spaces and routers (implicit broadcast).
\item \textbf{B4:} Explicit global workspace for arbitrary content routing.
\end{itemize}

\paragraph{Temporal depth (binding).}
Long-horizon tasks tempt departures from R3:
\begin{itemize}[leftmargin=1.2em,itemsep=0.2em]
\item \textbf{T1:} Atomic episodes.
\item \textbf{T2:} Short sequences for specific tasks.
\item \textbf{T3:} Extended narratives for planning.
\item \textbf{T4:} Autobiographical memory with causal models.
\end{itemize}

\paragraph{Learning sophistication (credit assignment).}
Capability pressure may violate R4:
\begin{itemize}[leftmargin=1.2em,itemsep=0.2em]
\item \textbf{L0:} Frozen deployment, possibly no training (innate Q1 witness baseline).
\item \textbf{L1:} Local parameter tuning within $\mathcal{S}$ (disjoint optimizer).
\item \textbf{L2:} Multiple $\mathcal{S}$ submaps adapt (still module-local).
\item \textbf{L3:} Cross-module credit ($\mathcal{S}$$~\leftrightarrow$perception/memory).
\item \textbf{L4:} End-to-end optimization through all pathways.
\end{itemize}

\subsubsection{Implicit violations through learned representations}\label{sec:q3-implicit}

The gradual paths above (S1-S4, B1-B4, T1-T4, L1-L4) focus on explicit architectural changes. However, functionally equivalent R1--R4 violations may arise through \emph{implicit} emergence within learned components, even when explicit architecture remains fixed. Neural networks implementing the categorization function (A1), need-to-affect mapping (A2), or policy scoring (A4--A5) might develop representations that violate our constraints:

\begin{itemize}[leftmargin=1.25em,itemsep=0.25em]
\item \textbf{Implicit global broadcast (R1):} Shared encoder layers across modules may develop latent spaces that function as covert hubs, making content available to heterogeneous consumers without explicit broadcast channels.

\item \textbf{Implicit metarepresentation (R2):} A categorization network might learn embeddings that encode not just situational features but also the agent's own internal states, history, or identity---self-referential content emerging from training incentives rather than architectural design.

\item \textbf{Implicit autobiographical consolidation (R3):} Even with step-local keys, learned representations might encode trajectory information if training involved temporally extended prediction tasks.

\item \textbf{Implicit cross-module coupling (R4):} Pre-deployment training with shared objectives or end-to-end gradients may leave residual dependencies frozen into the weights, even if deployment satisfies gradient isolation.
\end{itemize}

\subsubsection{Risk posture and audits}\label{sec:q3-risk}

A cautious policy for deployments aiming to maintain separation witness status is to keep systems at self-modeling levels \textbf{S1/S2}, broadcast level \textbf{B1}, temporal level \textbf{T1}, and learning levels \textbf{L0-L2}. Moving toward levels 3/4 on any axis represents graded departure from the conservative region.

\begin{table}[ht!]
\centering
\scalebox{0.8}{%
\small
\begin{tabular}{@{}p{0.10\linewidth}p{0.26\linewidth}p{0.30\linewidth}p{0.24\linewidth}@{}}
\toprule
\textbf{Criterion} & \textbf{Design indicators} & \textbf{Explicit tests} & \textbf{Implicit tests} \\
\midrule
R1 & Max fan-out of internal states; shared buffers or workspace tokens; cross-module attention; shared context objects; stealth hubs via shared embeddings/gates/routers & Count consumer modules per signal; ablate shared structures and check multi-module failure; cap/log inter-module bandwidth; scan for global-memory tokens or attention making content concurrently readable by heterogeneous consumers & Probe for latent hub structure in learned representations; measure cross-module mutual information \\
R2 & Paths that re-encode internal states as inputs; self-referential features; introspection nodes; variables about internal states (self-reports, self-ratings) feeding back into scoring or learning & Trace whether system inputs include own state variables; telemetry one-way only; check for telemetry leakage (debug/UX strings or dashboards reused in training) & Train classifiers to decode self-referential content (identity, internal states) from activations \\
R3 & Effective temporal receptive field; cross-episode keys; persistent profiles; cross-episode profile vectors; sequence models over long histories; identity/time keys & Shuffle/trim histories and test decision invariance; forbid write-back of summaries into system-readable buffers; check for offline summarizers writing autobiographical state & Probe whether trajectory or autobiographical information is decodable from ostensibly step-local representations \\
R4 & Gradient paths crossing modules; shared optimizers; multi-agent distillation; end-to-end backprop across modules or actors; shared routers coupling local states & Verify zero cross-gradients at boundaries; one optimizer per module; intervene in A and test drift in B under frozen losses & Audit training-time objectives for integration incentives; check for residual cross-module dependencies in frozen weights \\
\bottomrule
\end{tabular}
}
\caption{Preliminary audit indicators for R1--R4. Explicit tests verify architectural compliance; implicit tests detect learned violations that may emerge despite architectural constraints.}
\label{tab:audit-checklist}
\end{table}

Table~\ref{tab:audit-checklist} sketches design indicators and tests for both explicit architectural violations and implicit violations arising within learned components. Explicit tests verify compliance at the architectural level; implicit tests target emergent properties through indirect probes such as activation clustering, representational similarity analysis, and causal tracing~\citep{kaestner2024mechanistic,sharkey2025open}. Implementations may align logged variables with indicator-style assessments and document processes in the spirit of AI-audit frameworks~\citep{Raji2020Audit,NIST2023AIRMF}. Detecting implicit violations in artificial neural networks shares some of the challenges of inferring functional organization from biological connectomics. Artificial systems offer instrumental advantages---arbitrary interventions and exact readout of any internal variable at arbitrary precision. Yet invasive probing (e.g., ablation-style interventions) for access consciousness in artificial systems that might possess it would raise ethical concerns (e.g., consent, welfare) analogous to those arising in biological research. Developing robust methods to reveal the design elements targeted by R1--R4---whether encoded in weights or arising dynamically through data-dependent activation---remains an important open problem for AI interpretability~\citep{sharkey2025open} and R1--R4 audits at scale.

\FloatBarrier

\section{Synthesis and Outlook}\label{sec:synth_outlook}

\paragraph{Summary of findings.}
Under our \emph{audit proxy} for access risk (operationalized via R1--R4: no global workspace-like broadcast, no metarepresentation, no autobiographical consolidation, bounded learning), we argued \textbf{Q1} in the affirmative: there exists a \emph{separation witness}---an  emotion-like controller incorporating architectural principles A1--A8 and satisfying R1--R4, thereby remaining below access-like thresholds \emph{according to this proxy}. Informally, the witness architecture is ``too simple to be conscious yet rich enough to be emotional.'' This slogan captures the core intuition but should not be taken too literally; what makes the stance precise and falsifiable is the concrete controller of Fig.~\ref{fig:control_flow} tested against R1--R4 audits. For \textbf{Q2}, we identified \emph{stable design pockets}: functionally useful modifications that preserve R1--R4 compliance and thus maintain witness status. For \textbf{Q3}, we sketched \emph{graded routes} out of those pockets, proposing preliminary audit criteria to track progressive departure from the conservative region.

\paragraph{Potential use in comparative biology.} The separation framework, if sound, may license hypotheses beyond artificial systems. One could ask whether simpler organisms implement something approximating A1--A8 while lacking the integrative and reflective capacities associated with access consciousness---capacities whose graded relaxation we traced in Q3. Indeed, the framework suggests an intuition pump: starting from humans, where emotional life and consciousness are typically entangled, and moving toward simpler animals that we may intuitively credit with emotional responses but with diminished reflective and integrative capacities, one may approach---in the limit---systems where emotion-like control operates without access-enabling features. Whether this picture is correct is an empirical question, but the framework would render it coherent. 

This perspective might bear on ongoing debates. Researchers studying insects often speak of ``internal states'' or ``behavioral modulation'' rather than ``affect'' or ``emotion,'' and whether such phenomena merit the stronger label remains contested~\citep{anderson2014framework,adolphs2018structure}. Our architectural specification (A1--A8) sets a threshold for warranting that label in our sense, one that may apply even to primitive organisms: hierarchical organization with categorical abstraction, dual-source modulation through need appraisal and episodic memory, and affect-tagged storage enabling experience-dependent behavioral change. Below this threshold would lie merely reactive control. The Q1 witness shows that also above this threshold systems need not, as a matter of architectural necessity, possess the access-enabling features targeted by R1--R4. Whether particular biological organisms implement something approximating A1--A8 is an empirical question; our contribution, if accepted, would support the conditional claim that \emph{if} they do, this alone would not imply access-like consciousness under the proxy adopted here. The witness might thus serve double duty: as an existence proof for artificial systems, and as a scaffold for hypotheses about which functional details of affect-like control could have emerged evolutionarily as an intermediate layer between reactive regulation and conscious access~\citep{panksepp2012archaeology}.

\paragraph{Generalizability of the method.}

The present account could be turned into a blueprint that generalizes beyond our specific choices. The core method for identifying a potential separation witness (Q1) consists in: (i) implementing a candidate architecture (here A1--A8) in deliberately primitive form (here an architecturally frozen version of Fig.~\ref{fig:control_flow} with limited interfaces and strictly constrained read/write operations; see~\S\,\ref{sec:Q1}), and (ii) auditing it against theory-derived risk-reduction criteria (here R1--R4). This method can be applied to alternative architectural commitments $\text{A}_1-\text{A}_n$ and  risk-reduction sets $\text{R}_1-\text{R}_m$.

Our approach to the architectural features A1--A8 proposed here as sufficient for the label ``emotion-like control'' is intentionally minimalistic. This permits considerable primitiveness in the separation witness,\footnote{See in particular the example code referenced in SI, Part~I.} which may prompt skepticism: surely this cannot be \emph{all} there is to emotions? Readers who subscribe to functional accounts but find our architectural principles incomplete or misspecified---for example, in drawing insufficient distinctions among different types of emotions, affect, and feelings---may substitute their own architectural principles $\text{A}_1-\text{A}_n$ and provide alternative implementations. The methodological template remains applicable regardless.

Regarding the risk-reduction criteria R1--R4, our approach is theory-informed and deliberately conservative: these criteria reflect features associated with \emph{access-like} consciousness across prominent access-oriented frameworks. Satisfying R1--R4 is meant to place a system in a plausibly access-excluding region \emph{under this proxy}, but the criteria are not claimed to be necessary or sufficient for establishing the presence or absence of consciousness in general. Here too, readers may identify omissions---for instance, distinctions between awareness and consciousness that our criteria may fail to capture. Such concerns may motivate deriving alternative risk-reduction criteria $\text{R}_1-\text{R}_m$, unless the existing constraints can be shown to already preclude the relevant mechanisms. Extensions may also attempt to formulate optional criteria tailored to theories that are not primarily access-oriented, such as Integrated Information Theory (IIT). For example, one could add an R5 aimed at reducing large-scale causal integration---using tractable approximations or proxies for $\Phi$---by targeting the formation of a tightly integrated complex~\citep{Tononi2016IIT30}. The methodology permits extending the risk-reduction set when one wishes to track additional theoretical risk lenses.

Future work may thus build on this template, adapting both the ``emotion-like'' specification and ``consciousness-risk-reducing'' tests as theoretical understanding evolves. Corresponding adaptations to the modification analysis (Q2) and the gradual relaxation of risk reductions (Q3) would follow the same principles.

\paragraph{Toward a robust separation conjecture.}
A broader conjecture emerges: if modifications establishing architectural principles $\text{A}_1-\text{A}_n$ to satisfy risk reductions $\text{R}_1-\text{R}_m$ consistently yield affirmative answers to Q1--Q3 across a wide variety of emotion-controller proposals, this would constitute strong evidence that emotion-like control, at least for basic emotions, is logically independent of access consciousness. This reasoning assumes that the architectural principles $\text{A}_1-\text{A}_n$ are reduced to their functional minimum. Negative answers to Q1 follow trivially if a model intrinsically presupposes (access) consciousness.\footnote{For example, \citet{ledoux2017higher} extend HOT to include emotions and write ``[...] we define `fear' as the conscious feeling one has when in danger [...]. One implication of our view is that emotions can never be unconscious. Responses controlled by subcortical survival circuits that operate nonconsciously sometimes occur in conjunction with emotional feelings but are not emotions.'' In contrast, \citet{LeDoux2003CMN} writes ``Fear and fear learning have been dealt with here without addressing the conscious experience of fear [...]. While this is more a problem about consciousness than about emotion, [...] research on emotion may be able to contribute''. Under the assumption that adding access-enabling features weakly increases the probability of phenomenality, our methodology seems more readily applicable to this earlier framing.}  
Stipulative approaches settle the question by definition.\footnote{One might object that A1--A8 are themselves stipulative. We acknowledge this: any demarcation involves definitional choices. The crucial difference is that A1--A8 stipulate a \emph{floor}---a minimal threshold distinguishing emotion-like control from mere reactive systems---while leaving open whether additional capacities (such as access consciousness) are also necessary. Stipulating a ceiling (e.g., that emotions \emph{require} consciousness) forecloses the separation question by definition; stipulating a floor preserves it as empirically tractable.}
A genuine test arises when a model does \emph{not} presuppose access consciousness, yet the risk reductions $\text{R}_1-\text{R}_m$ prove difficult or impossible to satisfy---suggesting that access-enabling features may be necessary after all. Higher-order social emotions (e.g., guilt, shame, pride) that appear to require self-referential processing seem to be potential candidates for this outcome. Conversely, ease in satisfying $\text{R}_1-\text{R}_m$ provides preliminary evidence against the necessity of access consciousness. Different categories of emotions may thus yield different answers to Q1; the methodology accommodates this heterogeneity.

The present analysis represents a first step toward this broader program, offering initial results alongside a methodological template. For example, whether short-term guilt- or shame-equivalent states can be captured within our model (or slight extensions)---e.g., via observed cues and discrepancies in a need for bonding---becomes amenable to operational test. Dogs display guilt-like behaviors correlating with owner cues, casting doubt on whether genuine guilt is present~\citep{horowitz2009guilty}. An alternative interpretation is that received anger cues produce guilt-like uneasiness that recurs under similar circumstances, triggers a ``guilty look'' policy reinforced by bonding success, and operates without higher-order self-representation or autobiographical consolidation.
The emotion-like controller presented here could perhaps mirror such internal states and guide external behavior in sociable robots~\citep{Breazeal2003IJHCS} with minimal extensions (e.g., longer episode duration). This raises the possibility that even some primitive forms of guilt-like control are functionally implementable independently of whatever further capacities---if any---distinguish conscious from non-conscious affect. Whether and how strongly the dog may be said to \emph{feel} remains an open question on which theorists would presumably disagree~\citep{panksepp2012archaeology,ledoux2017higher}; our framework offers a functionalist approach testable in artificial systems as one basis for advancing such debates on conscious experience---now arising in analogous form for AI~\citep{butlin2025identifying,seth2025conscious}.

\vspace{0.5em}

\paragraph{Closing note.}
Whatever one's stance on consciousness or emotion, the present contribution may be engaged at \emph{three} independent levels.

\emph{(1) A practical engineering toolkit.}
Readers who disagree with our methodology may nonetheless find value in applying the concrete, modular controller based on principles A1--A8, with its intuitive tuning parameters (drives, valence, arousal). The dual-source design (immediate needs $+$ episodic hints) aims to add robustness; the hierarchical separation (situations $\!\to\!$ categories $\!\to\!$ policies $\!\to\!$ actions) is intended to manage complexity; and explicit interfaces (keys, tags, scores) are designed to make systems debuggable and ablatable.

\emph{(2) A bridge from theory to measurement.}
Questions such as ``What are emotions?'' and ``Can emotions exist without consciousness?'' are notoriously vague. We propose a systematic approach for making such questions tractable within a functionalist framework: specify architectural principles, implement them in auditable form, and test against theory-derived risk-reduction criteria. The resulting template---illustrated here through a separation witness, stability analysis, and risk gradients---may prove applicable to other questions concerning emotions and/or consciousness.

\emph{(3) A safety framework potentially transcending emotion architectures.}
Our risk-reduction criteria (R1--R4) are intended to function as \emph{access consciousness risk reduction audits}. They cannot certify the absence of consciousness, but they flag architectural patterns that multiple theories associate with access-consciousness risk---and by relaxing criteria successively, they enable tracing graded paths toward such risk. These theory-informed \emph{risk reduction proxies} are meant to delineate a plausibly access-excluding region \emph{under this proxy}: for explicit architectural features, they may offer directly actionable, falsifiable tests; for implicit violations emerging within learned representations, the proposed audits remain preliminary. In principle this methodology is not specific to the examined scenarios and it may find use in audits of other AI architectures.

We hope each community finds something of value here: engineers, an architecture to experiment with; theorists, a conceptual scaffold and a family of testable cases; policymakers and safety researchers, a suggestion for structuring audits through access consciousness risk-reduction criteria. As capabilities accelerate while theories of consciousness remain contested, definitive conclusions may lie beyond reach. Yet navigation tools may prove more tractable than definitive answers---grounding architectural choices in biologically inspired principles while keeping them transparent and auditable. The templates and tests proposed here are intended as starting points for discovery, not destinations, and are designed to evolve as understanding advances.




\section*{Declarations}

\begin{itemize}
\item \textbf{Published Article:} Supplementary Information and the published version is available online at \citep{Borotschnig2026Springer}. 
\item \textbf{Code availability:} Reference implementations are available online at \url{https://github.com/affect-based-control/synthetic-emotion-controller}
\end{itemize}


\end{document}